\documentclass[11pt,a4paper]{article}
\usepackage[hyperref]{eacl2021}
\usepackage{times}
\usepackage{latexsym}

\usepackage{microtype}
\usepackage{graphicx}
\usepackage{subcaption}
\usepackage{enumitem}
\usepackage{booktabs,subcaption,amsmath}
\usepackage{multirow}
\usepackage{makecell}
\usepackage{hyperref}

\usepackage{cleveref}
\crefname{section}{\S}{\S\S}
\Crefname{section}{\S}{\S\S}
\crefname{table}{Table.}{}
\crefname{figure}{Figure}{}
\crefname{algorithm}{Algorithm}{}
\crefname{equation}{equation}{}
\crefname{appendix}{Appendix}{}
\crefformat{section}{\S#2#1#3}  

\newcommand{\insertNewDataFlatTable}{
\begin{table*}[h!]
\begin{center}
\small
\begin{tabular}[b]{p{5cm}ccccccc}
\toprule
\multirow{2}{*}{\bf Model} & \bf en & \bf es & \bf fr & \bf de & \bf hi & \bf th & \bf{Avg}(5 langs)\\
& & & \multicolumn{4}{l}{(Exact Match Accuracy)} \\
\midrule
\addlinespace
\multicolumn{8}{l}{\textit{In-language models (only use target language training data)}} \\
\addlinespace
\midrule
XLU biLSTM & 78.2 & 70.8 & 68.9 & 65.1 & 62.6 & 68 & 67.1\\
XLM-R & 85.3 & 81.6 & 79.4 & 76.9 & 76.8 & 73.8 & 77.7\\
\addlinespace
\midrule
\addlinespace
\multicolumn{8}{l}{\textit{Multilingual models (use training data from multiple languages)}} \\
\addlinespace
\midrule
XLU biLSTM & 78.2 & 73.8 & 71.5 & 65.8 & 63.1 & 68.7 & 68.6\\
XLM-R & 86.3 & 83.6 & 81.8 & 79.2 & 78.9 & 76.7 & 80\\
\addlinespace
\midrule
\addlinespace
\multicolumn{8}{l}{\textit{Zero-shot target language models (only use English training data)}} \\
\addlinespace
\midrule
XLM-R on EN & N/A & 69.1 & 65.4 & 64 & 55 & 43.8 & 59.5\\
XLM-R with mask in \cref{sec:flat_mask} & N/A & 68 & 69.5 & \textbf{69.2} & \textbf{63.3} & 35.3 & 61.1\\
XLM-R on EN + translate align ~\cref{sec:translate_align} & N/A & 74.5 & \textbf{72.6} & 64.7 & 58.3 & \textbf{56.5} & 65.3\\
XLM-R with mask + translate align & N/A & \textbf{74.6} & 72.2 & 65.7 & 62.5 & 53.2 & \textbf{65.6}\\
\addlinespace
\bottomrule
\end{tabular}
\caption{Results on flat representation for 6 languages. We report exact match accuracy in this table. More metrics including intent accuracy and slot F1 is in Table \ref{tab:7} in Appendix. Notice that average is calculated across 5 languages except English to be comparable to zero-shot results. Best result for zero-shot is in bold. Taking best zero shot setting for each language, average exact match accuracy is 67.2. Note that for zero-shot setting, we only use EN train and eval data without any target language data.
\label{tab:1}}
\end{center}
\end{table*}
}

\newcommand{\insertNewDataDecoupledTable}{
\begin{table*}[h!]
\begin{center}
\small
\begin{tabular}[b]{p{6cm}ccccccc}
\toprule
\multirow{2}{*}{\bf Model} & \bf en & \bf es & \bf fr & \bf de & \bf hi & \bf th & \bf{Avg}(5 langs)\\
& & & \multicolumn{4}{l}{(Exact Match Accuracy)} \\
\midrule
\addlinespace
\multicolumn{8}{l}{\textit{In-language models (only use target language training data)}} \\
\addlinespace
\midrule
XLU biLSTM & 77.8 & 66.5 & 65.6 & 61.5 & 61.5 & 62.8 & 63.6\\
XLM-R encoder + random decoder & 83.9 & 76.9 & 74.7 & 71.2 & 70.2 & \textbf{71.2} & 72.8\\
mBART & 81.8 & 75.8 & 68.1 & 69.1 & 67.6 & 61.2 & 68.4\\
mBART on MT & \textbf{84.3} & 77.2 & 74.4 & 70.1 & 69.2 & 66.9 & 71.6 \\
CRISS & 84.2 & \textbf{78} & \textbf{75.5} & \textbf{72.2} & \textbf{73} & 68.8 & \textbf{73.5}\\
MARGE & 84 & 77.7 & 75.4 & 71.5 & 70.8 & 70.8 & 73.2\\
\addlinespace
\midrule
\addlinespace
\multicolumn{8}{l}{\textit{Multilingual models (use training data from multiple languages)}} \\
\addlinespace
\midrule
XLM-R encoder + random decoder & 83.6 & \textbf{79.8} & \textbf{78} & 74 & 74 & \textbf{73.4} & \textbf{75.8}\\
mBART & 83 & 78.9 & 76 & 72.9 & 72.8 & 68.8 & 73.9\\
CRISS & \textbf{84.1} & 79.1 & 77.7 & \textbf{74.4} & \textbf{74.7} & 71.3 & 75.4\\
\addlinespace
\midrule
\addlinespace
\multicolumn{8}{l}{\textit{Zero-shot target language models (only use English training data)}} \\
\addlinespace
\midrule
XLM-R on EN & N/A & 50.3 & 43.9 & 42.3 & 30.9 & 26.7 & 38.8\\
XLM-R on EN + translate align & N/A & 71.9 & 70.3 & 62.4 & 63 & \textbf{60} & \textbf{65.5}\\
CRISS on EN & N/A & 48.6 & 46.6 & 36.1 & 31.2 & 0 & 32.5\\
CRISS on EN + translate align & N/A & \textbf{73.3} & \textbf{71.7} & \textbf{62.8} & \textbf{63.2} & 53 & 64.8\\
\addlinespace
\bottomrule
\end{tabular}
\caption{Results on compositional decoupled representation for 6 languages. Metric is exact match accuracy. Average is calculated across 5 languages except English. Best result for each setting is in bold. For reference, exact match accuracy for BART model in-language training for en is 84.6.
\label{tab:2}}
\end{center}
\end{table*}
}

\newcommand{\insertAtisAndTopTable}{
\begin{table*}[h!]
\begin{center}
\small

\begin{tabular}[b]{p{4cm}|cc|cc}
\toprule
\multirow{2}{*}{\bf Model} & \multicolumn{2}{c|}{\bf Multilingual ATIS} & \multicolumn{2}{c}{ \bf Multilingual TOP} \\
& \bf hi & \bf tr & \bf es & \bf th \\
\midrule
\addlinespace
\multicolumn{5}{l}{\textit{In-language models (only use target language training data)}} \\
\addlinespace
\midrule
Original paper & -/-/74.6 & -/-/75.5 & 74.8/96.6/83.0 & 84.8/96.6/90.6 \\
XLM-R & 53.6/80.6/84.4 & 52.6/90.0/80.4 & 84.3/98.9/90.2 & 90.6/97.4/95 \\
\addlinespace
\midrule
\addlinespace
\multicolumn{5}{l}{\textit{Multilingual models (use training data from multiple languages)}} \\
\addlinespace
\midrule
original paper (bilingual) & -/-/80.6 & -/-/78.9 & 76.0/97.5/83.4 & 86.1/96.9/91.5 \\
XLM-R ALL & 62.3/85.9/87.8 & 65.7/92.7/86.5 & 83.9/99.1/90 & 91.2/97.7/95.4 \\
\addlinespace
\midrule
\addlinespace
\multicolumn{5}{l}{\textit{Zero-shot target language models (only use English training data)}} \\
\addlinespace
\midrule
Original paper & N/A & N/A & 55/85.4/72.9 & \textbf{45.6/95.9/55.4} \\
MBERT MLT & N/A & N/A & -/87.9/73.9 & -/73.46/27.1 \\
XLM-R on EN & 40.3/80.2/76.2 & 15.7/78/51.8 & \textbf{79.9/97.7/84.2} & 35/90.4/46  \\
XLM-R with mask & 49.4/85.3/84.2 & 19.7/79.7/60.6 & 76.9/98.1/85 & 23.5/95.9/30.2 \\
XLM-R EN + translate align & 53.2/85.3/84.2 & \textbf{49.7/91.3/80.2} & 66.5/98.2/75.8 & 43.4/97.3/52.8 \\
XLM-R mask + translate align & \textbf{55.3/85.8/84.7} & 46.4/89.7/79.5 & 73.2/98/83 & 41.2/96.9/52.8 \\
\addlinespace
\bottomrule
\end{tabular}

\caption{Results on Multilingual ATIS and Multilingual TOP, metrics are exact match accuracy / intent accuracy / slot F1 respectively. For zero-shot, first line is from original dataset paper. Best result for zero-shot is in bold.
\label{tab:3}}
\end{center}
\end{table*}
}

\newcommand{\insertDatasetSummary}{
\begin{table*}[h!]
\begin{center}
\small

\begin{tabular}[b]{p{3cm}|cccccc|cc}
\toprule
\multirow{2}{*}{\bf Domain} & \multicolumn{6}{ c| }{\bf Number of utterances (training/validation/testing)} & \bf Intent & \bf Slot \\
& \bf English & \bf German & \bf French & \bf Spanish & \bf Hindi & \bf Thai & \bf types & \bf types \\
\midrule
\bf Alarm & 2,006 & 1,783 & 1,581 & 1,706 & 1,374 & 1,510 & 6 & 5 \\
\bf Calling & 3,129 & 2,872 & 2,797 & 2,057 & 2,515 & 2,490 & 19 & 14 \\
\bf Event & 1,249 & 1,081 & 1,050 & 1,115 & 911 & 988 & 12 & 12 \\
\bf Messaging & 1,682 & 1,053 & 1,239 & 1,335 & 1,163 & 1,082 & 7 & 15 \\
\bf Music & 1,929 & 1,648 & 1,499 & 1,312 & 1,508 & 1,418 & 27 & 12 \\
\bf News & 1,682 & 1,393 & 905 & 1,052 & 1,126 & 930 & 3 & 6 \\
\bf People & 1,768 & 1,449 & 1,392 & 763 & 1,408 & 1,168 & 17 & 16 \\
\bf Recipes & 1,845 & 1,586 & 1,002 & 762 & 1,378 & 929 & 3 & 18 \\
\bf Reminder & 1,929 & 2,439 & 2,321 & 2,202 & 1,781 & 1,833 & 19 & 17 \\
\bf Timer & 1,488 & 1,358 & 1,013 & 1,165 & 1,152 & 1,047 & 9 & 5 \\
\bf Weather & 2,372 & 2,126 & 1,785 & 1,990 & 1,815 & 1,800 & 4 & 4 \\
\midrule
\bf Total & 22,288 & 18,788 & 16,584 & 15,459 & 16,131 & 15,195 & 117 & 78 \\
\bottomrule
\end{tabular}
\caption{Summary statistics of the MTOP dataset. The Data is roughly divided into 70:10:20 percent splits for train, eval and test.
\label{table:4}}
\end{center}
\end{table*}
}

\newcommand{\insertFlatMoreResults}{
\begin{table*}[h!]
\begin{center}
\small
\begin{tabular}[b]{p{5cm}cccccc}
\toprule
\multirow{2}{*}{\bf Model} & \bf en & \bf es & \bf fr & \bf de & \bf hi & \bf th \\
& & & \multicolumn{4}{l}{(Intent Accuracy / Slot F1)} \\
\midrule
\addlinespace
\multicolumn{7}{l}{\textit{In-language models (only use target language training data)}} \\
\addlinespace
\midrule
XLU biLSTM & 94.0/88.6 & 90.1/83.0 & 89.6/81.8 & 88.8/81.4 & 85.9/79.6 & 91.2/80.4 \\
XLM-R & 96.7/92.8 & 95.2/89.9 & 94.8/88.3 & 95.7/88.0 & 94.4/87.5 & 93.4/85.4 \\
\addlinespace
\midrule
\addlinespace
\multicolumn{7}{l}{\textit{Multilingual models (use training data from multiple languages)}} \\
\addlinespace
\midrule
XLU biLSTM & 94.6/88.4 & 91.3/84.6 & 91.3/83.0 & 90.3/81.2 & 87.6/78.9 & 91.9/80.5 \\
XLM-R & 97.1/93.2 & 96.6/90.8 & 96.3/89.4 & 96.7/88.8 & 95.4/88.4 & 95.1/86.3 \\
\addlinespace
\midrule
\addlinespace
\multicolumn{7}{l}{\textit{Zero-shot target language models (only use English training data)}} \\
\addlinespace
\midrule
XLM-R on EN & N/A & 93.5/81.7 & 90.7/81.6 & 91.2/78.7 & 88.4/71.8 & 88.0/63.3 \\
XLM-R with mask in \cref{sec:flat_mask} & N/A & 94.7/81.0 & 93.9/82.0 & \textbf{94.0/81.8} & \textbf{94.1/77.3} & 92.0/56.4 \\
XLM-R on EN + translate align ~\cref{sec:translate_align} & N/A & 96.2/84.6 & \textbf{95.4/82.7} & 96.1/78.9 & 94.7/72.7 & \textbf{92.7/70.0} \\
XLM-R with mask + translate align & N/A & \textbf{96.3/84.8} & 95.1/82.5 & 94.8/80.0 & 94.2/76.5 & 92.1/65.6 \\
\addlinespace
\bottomrule
\end{tabular}
\caption{Intent Accuracy / Slot F1 for models in Table \ref{tab:1}.
\label{tab:7}}
\end{center}
\end{table*}
}
\newcommand{\insertENExample}{
\begin{figure}[t]
\begin{center}
\includegraphics[height=10cm]{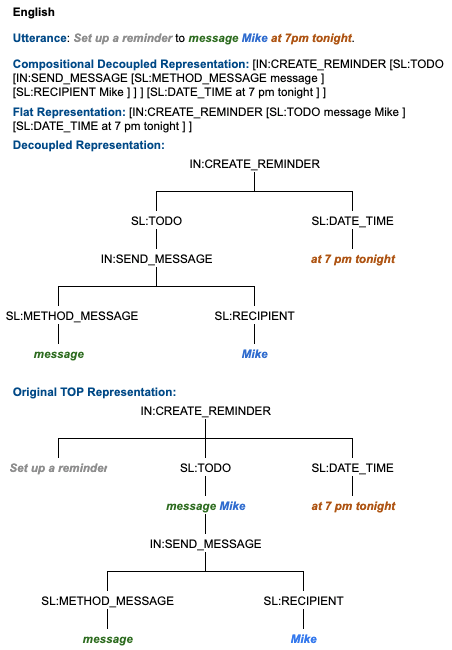}
\end{center}
\caption{
An English example from the data, showing its \textit{flat} representation and \textit{compositional decoupled} representation and a comparison between the decoupled and the original TOP representations in tree format.
}
\label{fig:en_example}
\end{figure}
}

\newcommand{\insertDEExample}{
\begin{figure}[t]
\begin{center}
\includegraphics[height=6cm]{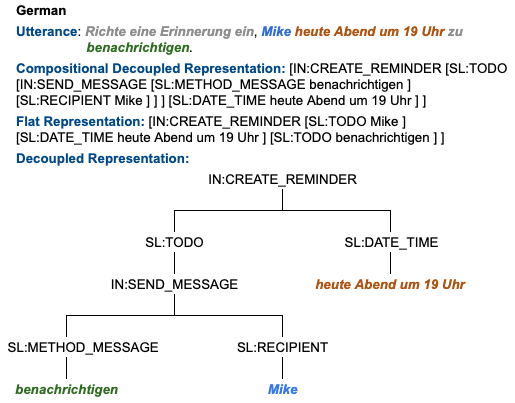}
\end{center}
\caption{
German utterance constructed from the English example of Figure~\ref{fig:en_example}. Even though the slot text order changed, we can still easily build a decoupled representation with the same structure. 
}
\label{fig:de_example}
\end{figure}
}

\aclfinalcopy 

\newcommand\blankfootnote[1]{%
  \let\thefootnote\relax\footnotetext{#1}%
  \let\thefootnote\svthefootnote%
}
\title{MTOP: A Comprehensive Multilingual Task-Oriented Semantic Parsing Benchmark}

\author{
  Haoran Li \quad Abhinav Arora \quad Shuohui Chen \quad Anchit Gupta \\
  \\
  \textbf{Sonal Gupta \quad Yashar Mehdad} \\
  \\
  Facebook \\
}
  
\date{}

\begin{document}
\maketitle
\blankfootnote{Correspondence to \texttt{\{aimeeli,abhinavarora\} @fb.com}}
\begin{abstract}
Scaling semantic parsing models for task-oriented dialog systems to new languages is often expensive and time-consuming due to the lack of available datasets. Available datasets suffer from several shortcomings: a) they contain few languages b) they contain small amounts of labeled examples per language c) they are based on the simple intent and slot detection paradigm for non-compositional queries. In this paper, we present a new multilingual dataset, called \textbf{MTOP}, comprising of 100k annotated utterances in 6 languages across 11 domains. We use this dataset and other publicly available datasets to conduct a comprehensive benchmarking study on using various state-of-the-art multilingual pre-trained models for task-oriented semantic parsing. We achieve an average improvement of +6.3 points on Slot F1 for the two existing multilingual datasets, over best results reported in their experiments. Furthermore, we demonstrate strong zero-shot performance using pre-trained models combined with automatic translation and alignment, and a proposed distant supervision method to reduce the noise in slot label projection. 
\end{abstract}

\section{Introduction}

With the rising adoption of virtual assistant products, task-oriented dialog systems have been attracting more attention in both academic and industrial communities. One of the first steps in these systems is to extract meaning from the natural language used in conversation to build a semantic representation of the user utterance. Typical systems achieve this by classifying the \textit{intent} of the utterance and tagging the corresponding \textit{slots}. With the goal of handling more complex queries, recent approaches propose hierarchical representations~\citep{gupta2018semantic} that are expressive enough to capture the task-specific semantics of complex nested queries.


Although, there have been sizable efforts around developing successful semantic parsing models for task-oriented dialog systems in English \citep{mesnil2013investigation, LiuL16, gupta2018semantic, rongali2020don}, we have only seen limited works for other languages. This is mainly due to the painstaking process of manually annotating and creating large datasets for this task in new languages. In addition to the shortage of such datasets, existing datasets \citep{UpadhyayFTHH18, schuster2018cross} are not sufficiently diversified in terms of languages and domains, and do not capture complex nested queries. This makes it difficult to perform more systematic and rigorous experimentation and evaluation for this task across multiple languages.

Building on these considerations and recent advancements on cross-lingual pre-trained models \citep{devlin-etal-2019-bert, lample2019cross, xlmr}, this paper is making an effort to bridge the above mentioned gaps.  The main contributions of this paper can be summarized as follows:
\begin{itemize}
\item MTOP Dataset: We release an almost-parallel multilingual task-oriented semantic parsing dataset covering \textbf{6 languages} and \textbf{11 domains}. To the best of our knowledge, this is the first multilingual dataset which contains compositional representations that allow complex nested queries.
\item We build strong benchmarks on the released MTOP dataset using state-of-the-art multilingual pre-trained models for both flat and compositional representations. We demonstrate the effectiveness of our approaches by achieving new state-of-the-art result on existing multilingual task-oriented semantic parsing datasets.
\item We demonstrate strong performance on zero-shot cross-lingual transfer using automatic translation and alignment, combined with a proposed distant supervision approach. We achieve 67.2\% exact match accuracy (averaged across 5 languages) without using any target language data compared to best in-language model performance of 77.7\%. 
\end{itemize}

\section{Related Work}

\paragraph{Task-Oriented Semantic Parsing} 
The majority of the work on task-oriented dialog systems has been centered around intent detection and slot filling - for example, the representations used on the ATIS dataset \citep{mesnil2013investigation, LiuL16, ZhuY17} and in the Dialog State Tracking Challenge \citep{WilliamsRH16a}. This essentially boils down to a text classification and a sequence labeling task, which works great for simple non-compositional queries. For more complex queries with recursive slots, state of the art systems use hierarchical representations, such as the TOP representation \cite{gupta2018semantic}, that is modeled using Recurrent Neural Network Grammars \cite{DyerKBS16} or as a Sequence to Sequence task \citep{rongali2020don}.

\paragraph{Pre-trained Cross-lingual Representation}
Over the past few years, pre-trained cross-lingual representations have demonstrated tremendous success in achieving state of the art in various NLP tasks. The majority of the earlier work focuses on cross-lingual emebedding alignment~\citep{MikolovLS13, ammar2016massively, LampleCRDJ18}. \citet{Schuster2019cross} further extend upon this by aligning contextual word embeddings from the ELMo model \citep{Peters2018}. Later with the success of Transformer~\citep{vaswani2017attention} based masked language model pre-training, \citet{devlin-etal-2019-bert} and \citet{lample2019cross} introduce mBERT and XLM respectively, and \citet{PiresSG19} show the effectiveness of these on sequence labeling tasks. \citet{xlmr} present XLM-R, a pre-trained multilingual masked language model trained on data in 100 languages, that provides strong gains over XLM and mBERT on classification and sequence labeling tasks. 

The models discussed above are encoder-only models. More recently, multilingual seq-to-seq pre-training has become popular. \citet{journals/corr/abs-2001-08210} introduce mBART, a seq-to-seq denoising auto-encoder pre-trained on monolingual corpora in many languages, which extends BART \citep{lewis2019bart} to a multilingual setting. More recently, \citet{lewis2020pre} introduced a seq-to-seq model pre-trained on a multilingual multi-document paraphrasing objective, which self-supervises the reconstruction of target text by retrieving a set of related texts and conditions on them to maximize the likelihood of generating the original. \citet{tran2020cross} is another contemporary work that mines parallel data using encoder representations and jointly trains a seq-to-seq model on this parallel data.
\paragraph{Cross-Lingual Task-Oriented Semantic Parsing} 
Due to the ubiquity of digital assistants, the task of cross-lingual and multilingual task-oriented dialog has garnered a lot of attention recenty, and few multilingual benchmark datasets have been released for the same. To the best of our knowledge, all of them only contain simple non-compositional utterances, suitable for the intent and slots detection tasks. \citet{UpadhyayFTHH18} release a benchmark dataset in Turkish and Hindi (600 training examples), obtained by translating utterances from the ATIS corpus \citep{Price90} and using Amazon Mechanical Turk to generate phrase level slot annotation on translations. \citet{schuster2018cross} release a bigger multilingual dataset for task-oriented dialog in English, Spanish and Thai across 3 domains. They also propose various modeling techniques such as using XLU embeddings (see \citet{ruder2017survey} for literature review) for cross-lingual transfer, translate-train and ELMo \citep{Peters2018} for target language training. BERT-style multilingual pre-trained models have also been applied to task-oriented semantic parsing. \citet{castellucci2019multi} use multilingual BERT for joint intent classification and slot filling, but they don't evaluate on existing multilingual benchmarks. Instead, they introduce a new Italian dataset obtained via automatic machine translation of SNIPS \citep{snips}, which is of lower quality. For zero shot transfer, \citet{aaai/LiuWLXF20} study the idea of selecting some parallel word pairs to generate code-switching sentences for learning the inter-lingual semantics across languages and compare the performance using various cross-lingual pre-trained models including mBERT and XLM.

\insertDatasetSummary

\section{Data} \label{sec:data} 

Existing multilingual task-oriented dialog datasets, such as ~\citet{UpadhyayFTHH18, schuster2018cross}, rely on expensive manual work for preparing guidelines and annotations for other languages; which is probably why they only contain very few languages and few labeled data examples for other languages. Furthermore, annotations will be more complicated and expensive if they were to include compositional queries, where slots can have nested intents. To this end we create an almost parallel multilingual task-oriented semantic parsing corpora which contains \textbf{100k examples} in total for \textbf{6 languages} (both high and low resource): \textit{English}, \textit{Spanish}, \textit{French}, \textit{German}, \textit{Hindi} and \textit{Thai}. Our dataset contains a mix of both simple and compositional nested queries across \textbf{11 domains}, \textbf{117 intents} and \textbf{78 slots}. \cref{table:4} shows a summary statistics of our MTOP dataset.

We release the dataset at \url{https://fb.me/mtop_dataset}.

\subsection{Dataset Creation}
Our approach for creating this dataset consists of two main steps: i) generating synthetic utterances and annotating in English, ii) translation, label transfer, post-processing, post editing and filtering for other 5 languages. Generating the English utterances and their annotations, for the 11 domains, follows the exact process as described in \cite{gupta2018semantic}. We ask crowdsourced workers to generate natural language sentences that they would ask a system which could assist in queries corresponding to our chosen domains. These queries are labeled by two annotators. A third annotator is used only to adjudicate any disagreements. Once an annotated English dataset is available, we build the multilingual dataset through the following steps:

\paragraph{Translation:} We first extract slot text spans from English annotation and present the utterances along with slot text spans to professional translators for translation to the target language. We prepare detailed guidelines, where we ask the translators to ensure that the translation for each slot span is exactly in the same way as it occurs in the translated utterance. For example, when translating the slot span \textit{mom} in utterance \textit{call my mom}, we ask the translators to use the same target language word for \textit{mom}, that they used in the translation for \textit{call my mom}.

\paragraph{Post-processing:} After we obtain the translation of utterances and corresponding slot text spans, we use the tree structure of English and fill in the translated slot text spans to construct the annotation in the target languages. Our representation, described in \cref{sec:decoupled}, enables us to reconstruct the annotations.

\paragraph{Post-editing and Quality Control:} We further run two rounds of quality control over translated utterances and slots, and revise the data accordingly. In the first round, we ask translators to review and post-edit the errors in translations and slot alignments. In the second round, the constructed target language data is presented to different annotators for a lightweight annotation quality review. 83\% of the data was marked as good quality data and passed our quality standards, which can be interpreted as the inter-annotator agreement rate on the translated data. Based on this feedback, we remove low quality annotations from the dataset.

To create this dataset, for each target language we had three translators: two were responsible for translation and the third one for review and edits. All the translators were professional translators, with native or close to native speaker skills. The overall time spent was 15 to 25 days for each language. Even though we run rigorous quality control, a dataset built by translation is bound to have few errors, such as using words or phrases that are not commonly used in spoken language. 

\insertENExample
\insertDEExample

\subsection{Data Format}
In this dataset, we release two kinds of representations, which we refer to as \textit{flat} representations and \textit{compositional decoupled} representations, that are illustrated in Figure~\ref{fig:en_example} for an English utterance. Most existing annotations for task-oriented dialog systems follow the intent classification and slot tagging paradigm, which is what we refer to as the flat representation. Since our data contains compositional utterances with nested slots with intents within them, flat representations are constructed by only using the top level slots. We include the flat representation so that the data and the discussed modeling techniques are comparable to other task-oriented dialog benchmarks. To ensure the reproducibility of our results, we also release the tokenized version of utterances obtained via our in-house multilingual tokenizer. 

\subsubsection{Compositional Decoupled Representation} \label{sec:decoupled}
\citet{gupta2018semantic} demonstrate the inability of flat representations to parse complex compositional requests and propose a hierarchical annotation scheme (TOP representation) for semantic parsing, that allows the representation of such nested queries. We further use a representation, called the decoupled representation, that removes all the text from the TOP representation that does not appear in a leaf slot, assuming this text does not contribute to the semantics of the query. Figure~\ref{fig:en_example} highlights the difference between this decoupled representation and the original TOP representation. The decoupled representation makes the semantic representation more flexible and allows long-distance dependencies within the representation. It also makes translation-based data creation approach feasible for different languages despite syntactic differences, as the representation is \textit{decoupled} from the word order of the utterance. For example, in the German translation of the English example as shown in Figure~\ref{fig:de_example}, translations of \textit{message} and \textit{Mike} were separated by other words between them. However, it is straight forward to construct a decoupled representation as the representation is not bound by a word-order constraint.

\section{Model Architecture}
\label{sec:length}

\subsection{Joint intent and slot tagging for flat representation}
For flat representation, where there is a single top-level intent, the traditional way is to model it as an intent classification and a slot tagging problem. Our baseline model is a bidirectional LSTM intent slot model as described in \citet{LiuL16,ZhangW16a} with pre-trained XLU embeddings. Since existing pre-trained XLU embeddings (e.g., MUSE~\citep{LampleCRDJ18}) don't provide embedding for Hindi and Thai, we train our own using multiCCA following \citet{ammar2016massively}.
Compared to previous state-of-the-art work on existing multilingual task-oriented parsing datasets~\citep{aaai/LiuWLXF20, castellucci2019multi} which use Multilingual BERT, we use XLM-R~\citep{xlmr} since it's shown to outperform Multilingual BERT in cross-lingual performance on a variety of tasks. Specifically we use XLM-R Large in all our experiments. We use the same model architecture as in \citet{chen2019bert} and replace BERT encoder with XLM-R encoder. 

\subsection{Seq-to-seq for hierarchical representation}

Even though there are few existing works on cross lingual transfer learning for parsing flat representations, to the best of our knowledge, we are not aware of any other work that studies cross-lingual transfer for parsing complex queries in task-oriented dialog. In this section, we outline our modeling approaches for the compositional decoupled representation discussed in \cref{sec:decoupled}.

\paragraph{Seq-to-seq with Pointer-generator Network}
Our model adopts an architecture similar to \citet{rongali2020don}, where source is the utterance and target is the compositional decoupled representation described in \cref{sec:decoupled}. Given a source utterance, let $[\boldsymbol{e}_1,\boldsymbol{e}_2, ..., \boldsymbol{e}_n]$ be the encoder hidden states and $[\boldsymbol{d}_1,\boldsymbol{d}_2, ..., \boldsymbol{d}_m]$ be the corresponding decoder hidden states. At decoding time step t, the model can either generate an element from the ontology with generation distribution $\boldsymbol{p}^\text{g}_t$, or copy a token from the source sequence with copy distribution $\boldsymbol{p}^\text{c}_t$. Generation distribution is computed as:
$$
  \boldsymbol{p}^\text{g}_t = \operatorname{softmax}\left(\operatorname{Linear}_\text{g}[\boldsymbol{d}_t]\right)
$$
Copy distribution is computed as:
$$
\boldsymbol{p}^\text{c}_t, \boldsymbol{\omega}_t = \operatorname{MHA}\left(\boldsymbol{e}_1, ..., \boldsymbol{e}_n; \operatorname{Linear}_\text{c}[\boldsymbol{d}_t]  \right)
$$
where $\operatorname{MHA}$ stands for Multi-Head Attention~\citep{vaswani2017attention} and $\boldsymbol{\omega}_t$ is the attended vector used to compute the weight of copying $\boldsymbol{p}^\text{w}_t$:
$$
p^\text{w}_t = \operatorname{sigmoid}\left( \operatorname{Linear}_\alpha\left[\boldsymbol{d}_t ; \boldsymbol{\omega}_t\right] \right)
$$
The final probability distribution is computed as a mixture of the generation and copy distributions:
$$
\boldsymbol{p}_t = p^\text{w}_t \cdot \boldsymbol{p}^\text{g}_t + \left(1-p^\text{w}_t\right) \cdot \boldsymbol{p}^\text{c}_t.
$$
As a baseline, we use a standard LSTM encoder-decoder architecture with XLU embeddings. We also experiment with various transformer-based state of the art multilingual pre-trained models to improve upon the baseline. We use both pre-trained encoder-only models as well as pre-trained seq-to-seq encoder and decoder models. Here we outline the different models that we experimented with:




\begin{itemize}[leftmargin=*]
\setlength\itemsep{0em}
\item \textbf{XLM-R} encoder, pre-trained with masked language model objective in 100 languages. For decoder, we use randomly initialized transformer decoder as in \citet{vaswani2017attention}. 
\item \textbf{mBART}~\citep{journals/corr/abs-2001-08210} is pre-trained seq-to-seq model using denoising autoencoder objective on monolingual corpora in 25 languages.
\item \textbf{mBART on MT}: Machine translation is another common task for pre-training multilingual models. We follow \citet{tang2020multilingual} to further fine-tune mBART on English to 25 languages translation task. 
\item \textbf{CRISS}~\citep{tran2020cross} is pre-trained on parallel data in an unsupervised fashion. It iteratively mines parallel data using its own encoder outputs and trains a seq-to-seq model on the parallel data. CRISS has been shown to perform well on sentence retrieval and translation tasks. 
\item \textbf{MARGE}~\citep{lewis2020pre} is learned with an unsupervised multi-lingual multi-document paraphrasing objective. It retrieves a set of related texts in many languages and conditions on them to maximize the likelihood of generating the original text. MARGE has shown to outperform other models on a variety of multilingual benchmarks including document translation and summarization.
\end{itemize}

\section{Experiments}

We conduct thorough experiments\blankfootnote{We provide reproducibility details and all hyperparameters in \cref{sec:appendix}} on the new dataset we describe in in \cref{sec:data}. To further demonstrate the effectiveness of our proposed approaches, we also run additional experiments on the existing multilingual task-oriented semantic parsing datasets including \textit{Multilingual ATIS}~\citep{UpadhyayFTHH18} and \textit{Multilingual TOP}~\citep{schuster2018cross}. Note that both these data sets only include flat representation, while our data set contains hierarchical representations.

\subsection{Experimental Settings}
For all benchmarks, we have three different evaluation settings:
\begin{itemize}[leftmargin=*]
\setlength\itemsep{0em}
\item \textsc{In-language models}: We only use target language training data.
\item \textsc{Multilingual models}: We use training data in all available languages and train a single model for multiple languages.
\item \textsc{Zero-shot target language models}: We only use English data during training.
\end{itemize}
Next in each subsection we talk about details of approaches we use in these experiments.
\subsubsection{Translate and Align} \label{sec:translate_align}
With zero or few target language annotated examples, \textit{translate-train} is a common approach to augment target language training data. For semantic parsing tasks, besides translation we need alignment to project slot annotations to target language. This process is similar to how we collect our dataset, but using machine translation and alignment methods. For translation, we use our in-house machine translation system. We also tried other publicly available translation APIs and didn't find significant difference in final task performance. For alignment, we experimented with both, using attention weights from translation as in \citet{schuster2018cross} and fastalign~\citep{DBLP:conf/naacl/DyerCS13} and found data generated through fastalign leads to better task performance. Thus we only report results that use fastalign.

\subsubsection{Multilingual Training}
With the advancement of multilingual pre-trained models, a single model trained on multiple languages has shown to outperform in-language models~\citep{xlmr, hu2020xtreme}. As a result, we also experiment with multilingual training on our benchmark, including training jointly on all in-language data and training on English plus translated and aligned data in all other languages for the zero-shot setting. Instead of concatenating data in all languages together as in ~\citet{xlmr}, we adopt a multitask training approach where for each batch we sample from one language based on a given sampling ratio so that languages with fewer training data can be upsampled. We found this setting to perform better than mixed-language batches in our experiments.

\subsubsection{Distant Supervision in Zero-Shot Setting for Flat Representations} \label{sec:flat_mask}
Alignment models are not perfect, especially for low resource languages. To combat the noise and biases introduced in slot label projection, we experiment with another distant supervision approach in the zero-shot setting for learning flat representation models. We first concatenate the English utterance and its corresponding translation (using machine translation) in target language as input and then replace the English slot text with MASK token at random (30\% of the time, chosen empirically as a hyper-parameter). With the masked source utterance and the translated utterance as the concatenated input, we train a model to predict the overall intent and slot labels on the original English source. In this way, the MASK token can also attend to its translation counterpart to predict its label and the translated slot text could be distantly supervised by English labeled data.

\insertNewDataFlatTable{}
\section{Results and Discussions}
\subsection{Results on MTOP} \label{sec:mtopresult}

\paragraph{Flat Representation Results} \label{sec:flatformresult}
\cref{tab:1} shows the result on our MTOP dataset for all languages, using the flat representation. For both in-language and multilingual settings, XLM-R based models significantly outperform the BiLSTM models using XLU.
We also observe that multilingual models outperform in-language models.
Interestingly, for Hindi and Thai (both non-European languages), the improvements from multilingual training are considerably higher for XLM-R as compared to XLU BiLSTM. This observation highlights the remarkable cross-lingual transferability of the pre-trained XLM-R representations where fine-tuning on syntactically different languages also improves target language performance. 

For zero-shot cross-lingual transfer, we restrict ourselves to an XLM-R baseline to explore improvements using translate and align, and the distant supervision techniques as described in \ref{sec:translate_align} and \ref{sec:flat_mask} respectively. Our results demonstrate that distant supervision is able to considerably improve over the baselines for French, German and Hindi, while there is a small drop for Spanish. In the same setting, performance for Thai significantly degrades compared to the baseline. We suspect this is due to imperfect Thai tokenization that leads to learning noisy implicit alignments through distant supervision. The translate and align approach consistently improves over the baseline for all languages. It also performs better than distant supervision for all languages except German and Hindi. Our hypothesis is that the compounding nature of German inhibits the learning of hard alignment from fastalign. In summary, the XLM-R trained on all the 6 languages significantly outperforms all other models for this task.

In \cref{sec:more_results}, we further report intent accuracy and slot F1 metrics for the flat representation, as these are commonly used metrics in previous benchmarks for intent-slot prediction \citep{Price90, schuster2018cross}.

\insertNewDataDecoupledTable{}
\insertAtisAndTopTable{}
\paragraph{Compositional Decoupled Representation}
\cref{tab:2} shows the results on our MTOP dataset using compositional decoupled representation. In all settings, using multilingual pre-trained models significantly outperform the baseline. Surprisingly, mBART doesn't demonstrate strong performance compared to other models with fine-tuning on our task, even though fine-tuning BART on English achieves the best performance on English data. We hypothesize that mBART was under-trained for many languages and did not learn good cross-lingual alignments. In order to prove our hypothesis, we further fine-tune mBART on English to 25 languages translation task. The obtained mBART fine-tuned on translation significantly outperform the original mBART. The performance of CRISS and MARGE are at par with each other and among our best performing models across 5 languages, except Thai. XLM-R with random decoder performs the best on Thai. We believe this is because neither CRISS nor MARGE are pre-trained on Thai, while XLM-R pre-training includes Thai.

Similar to previous observations, multilingual training improves over the monolingual results. With multilingual training, XLM-R and CRISS are the best performing models for every language. Since XLM-R uses a randomly initialized decoder, it makes intuitive sense that such a decoder is better trained with multilingual training and thus obtains higher gains from more training data. Interestingly, mBART performance also improves a lot, which is another evidence that it was originally under-trained, as discussed in the previous paragraph. In the zero-shot setting, using the models fine-tuned on English does not perform well. In fact Thai zero shot using CRISS gives a 0 exact match accuracy, as the model was not pre-trained on any Thai data. Both XLM-R and CRISS show significant improvements when they utilized the machine translated and aligned data. 

\subsection{Results on Existing Benchmarks}
\cref{tab:3} shows results on two previously released multilingual datasets: Multilingual ATIS and Multilingual TOP. Similar to our findings in \ref{sec:flatformresult}, XLM-R based models significantly outperform the best results reported by the original papers and sets a new state-of-the-art on these benchmarks. Also, multilingual models trained on all available languages further improve the result. 

For Multilingual ATIS, in the zero-shot setting, our distant supervised masking strategy shows considerable gains compared to direct transfer using English. Using translate and aligned data also helps in improving the results significantly. When multitask trained together with masked data, it achieves the best zero-shot performance on Hindi. For both languages (Hindi and Turkish) this comes very close to the performance using target language training data. 

For multilingual TOP, direct transfer proves to be effective for Spanish, direct transfer from English overall yield better result than what's reported in Mixed-Language Training (MLT) with MBERT~\citep{aaai/LiuWLXF20}. While masking and translating generated data degrade its performance. Based on our error analysis, we find that tokenization mismatch, derived from translation data, causes such performance drop due to errors in slot text boundaries. For Thai, all our translation-based techniques perform worse than translate-train results from original paper. We attribute this primarily to the tokenization difference between our translated data and original test data. Unlike Spanish, Thai is much more sensitive to tokenization as it rarely uses whitespace.





\section{Conclusion}

In this paper, we release a new multilingual task-oriented semantic parsing dataset called \textbf{MTOP} that covers 6 languages, including both flat and compositional representations. We develop strong and comprehensive benchmarks for both representations using state-of-the-art multilingual pre-trained models in both zero-shot and with target language settings. We hope this dataset along with proposed methods benefit the research community in scaling task-oriented dialog systems to more languages effectively and efficiently.



\bibliography{anthology,eacl2021}

\begin{thebibliography}{37}
\expandafter\ifx\csname natexlab\endcsname\relax\def\natexlab#1{#1}\fi

\bibitem[{Ammar et~al.(2016)Ammar, Mulcaire, Tsvetkov, Lample, Dyer, and
  Smith}]{ammar2016massively}
Waleed Ammar, George Mulcaire, Yulia Tsvetkov, Guillaume Lample, Chris Dyer,
  and Noah~A Smith. 2016.
\newblock Massively multilingual word embeddings.
\newblock \emph{arXiv preprint arXiv:1602.01925}.

\bibitem[{Castellucci et~al.(2019)Castellucci, Bellomaria, Favalli, and
  Romagnoli}]{castellucci2019multi}
Giuseppe Castellucci, Valentina Bellomaria, Andrea Favalli, and Raniero
  Romagnoli. 2019.
\newblock Multi-lingual intent detection and slot filling in a joint bert-based
  model.
\newblock \emph{arXiv preprint arXiv:1907.02884}.

\bibitem[{Chen et~al.(2019)Chen, Zhuo, and Wang}]{chen2019bert}
Qian Chen, Zhu Zhuo, and Wen Wang. 2019.
\newblock Bert for joint intent classification and slot filling.
\newblock \emph{arXiv preprint arXiv:1902.10909}.

\bibitem[{Conneau et~al.(2020)Conneau, Khandelwal, Goyal, Chaudhary, Wenzek,
  Guzm{\'a}n, Grave, Ott, Zettlemoyer, and Stoyanov}]{xlmr}
Alexis Conneau, Kartikay Khandelwal, Naman Goyal, Vishrav Chaudhary, Guillaume
  Wenzek, Francisco Guzm{\'a}n, Edouard Grave, Myle Ott, Luke Zettlemoyer, and
  Veselin Stoyanov. 2020.
\newblock \href {https://doi.org/10.18653/v1/2020.acl-main.747} {Unsupervised
  cross-lingual representation learning at scale}.
\newblock In \emph{Proceedings of the 58th Annual Meeting of the Association
  for Computational Linguistics}, pages 8440--8451, Online. Association for
  Computational Linguistics.

\bibitem[{Coucke et~al.(2018)Coucke, Saade, Ball, Bluche, Caulier, Leroy,
  Doumouro, Gisselbrecht, Caltagirone, Lavril, Primet, and Dureau}]{snips}
Alice Coucke, Alaa Saade, Adrien Ball, Th{\'{e}}odore Bluche, Alexandre
  Caulier, David Leroy, Cl{\'{e}}ment Doumouro, Thibault Gisselbrecht,
  Francesco Caltagirone, Thibaut Lavril, Ma{\"{e}}l Primet, and Joseph Dureau.
  2018.
\newblock \href {http://arxiv.org/abs/1805.10190} {Snips voice platform: an
  embedded spoken language understanding system for private-by-design voice
  interfaces}.
\newblock \emph{CoRR}, abs/1805.10190.

\bibitem[{Devlin et~al.(2019)Devlin, Chang, Lee, and
  Toutanova}]{devlin-etal-2019-bert}
Jacob Devlin, Ming-Wei Chang, Kenton Lee, and Kristina Toutanova. 2019.
\newblock \href {https://doi.org/10.18653/v1/N19-1423} {{BERT}: Pre-training of
  deep bidirectional transformers for language understanding}.
\newblock In \emph{Proceedings of the 2019 Conference of the North {A}merican
  Chapter of the Association for Computational Linguistics: Human Language
  Technologies, Volume 1 (Long and Short Papers)}, pages 4171--4186,
  Minneapolis, Minnesota. Association for Computational Linguistics.

\bibitem[{Dyer et~al.(2013)Dyer, Chahuneau, and
  Smith}]{DBLP:conf/naacl/DyerCS13}
Chris Dyer, Victor Chahuneau, and Noah~A. Smith. 2013.
\newblock \href {https://www.aclweb.org/anthology/N13-1073/} {A simple, fast,
  and effective reparameterization of {IBM} model 2}.
\newblock In \emph{Human Language Technologies: Conference of the North
  American Chapter of the Association of Computational Linguistics,
  Proceedings, June 9-14, 2013, Westin Peachtree Plaza Hotel, Atlanta, Georgia,
  {USA}}, pages 644--648. The Association for Computational Linguistics.

\bibitem[{Dyer et~al.(2016)Dyer, Kuncoro, Ballesteros, and Smith}]{DyerKBS16}
Chris Dyer, Adhiguna Kuncoro, Miguel Ballesteros, and Noah~A. Smith. 2016.
\newblock \href
  {http://dblp.uni-trier.de/db/conf/naacl/naacl2016.html#DyerKBS16} {Recurrent
  neural network grammars.}
\newblock In \emph{HLT-NAACL}, pages 199--209. The Association for
  Computational Linguistics.

\bibitem[{Gupta et~al.(2018)Gupta, Shah, Mohit, Kumar, and
  Lewis}]{gupta2018semantic}
Sonal Gupta, Rushin Shah, Mrinal Mohit, Anuj Kumar, and Mike Lewis. 2018.
\newblock Semantic parsing for task oriented dialog using hierarchical
  representations.
\newblock In \emph{Proceedings of the 2018 Conference on Empirical Methods in
  Natural Language Processing}, pages 2787--2792.

\bibitem[{Hu et~al.(2020)Hu, Ruder, Siddhant, Neubig, Firat, and
  Johnson}]{hu2020xtreme}
Junjie Hu, Sebastian Ruder, Aditya Siddhant, Graham Neubig, Orhan Firat, and
  Melvin Johnson. 2020.
\newblock Xtreme: A massively multilingual multi-task benchmark for evaluating
  cross-lingual generalization.
\newblock \emph{arXiv preprint arXiv:2003.11080}.

\bibitem[{Izmailov et~al.(2018)Izmailov, Podoprikhin, Garipov, Vetrov, and
  Wilson}]{swa}
Pavel Izmailov, Dmitry Podoprikhin, Timur Garipov, Dmitry Vetrov, and
  Andrew~Gordon Wilson. 2018.
\newblock Averaging weights leads to wider optima and better generalization.
\newblock In \emph{Conference on Uncertainty in Artificial Intelligence (UAI
  2018)}.

\bibitem[{Kingma and Ba(2015)}]{kingma2014adam}
Diederik Kingma and Jimmy Ba. 2015.
\newblock Adam: A method for stochastic optimization.

\bibitem[{Lample and Conneau(2019)}]{lample2019cross}
Guillaume Lample and Alexis Conneau. 2019.
\newblock Cross-lingual language model pretraining.
\newblock \emph{Advances in Neural Information Processing Systems (NeurIPS)}.

\bibitem[{Lample et~al.(2018)Lample, Conneau, Ranzato, Denoyer, and
  Jégou}]{LampleCRDJ18}
Guillaume Lample, Alexis Conneau, Marc'Aurelio Ranzato, Ludovic Denoyer, and
  Hervé Jégou. 2018.
\newblock \href
  {http://dblp.uni-trier.de/db/conf/iclr/iclr2018.html#LampleCRDJ18} {Word
  translation without parallel data.}
\newblock In \emph{ICLR (Poster)}. OpenReview.net.

\bibitem[{Lewis et~al.(2020{\natexlab{a}})Lewis, Ghazvininejad, Ghosh,
  Aghajanyan, Wang, and Zettlemoyer}]{lewis2020pre}
Mike Lewis, Marjan Ghazvininejad, Gargi Ghosh, Armen Aghajanyan, Sida Wang, and
  Luke Zettlemoyer. 2020{\natexlab{a}}.
\newblock Pre-training via paraphrasing.
\newblock \emph{arXiv preprint arXiv:2006.15020}.

\bibitem[{Lewis et~al.(2020{\natexlab{b}})Lewis, Liu, Goyal, Ghazvininejad,
  Mohamed, Levy, Stoyanov, and Zettlemoyer}]{lewis2019bart}
Mike Lewis, Yinhan Liu, Naman Goyal, Marjan Ghazvininejad, Abdelrahman Mohamed,
  Omer Levy, Veselin Stoyanov, and Luke Zettlemoyer. 2020{\natexlab{b}}.
\newblock \href {https://doi.org/10.18653/v1/2020.acl-main.703} {{BART}:
  Denoising sequence-to-sequence pre-training for natural language generation,
  translation, and comprehension}.
\newblock In \emph{Proceedings of the 58th Annual Meeting of the Association
  for Computational Linguistics}, pages 7871--7880, Online. Association for
  Computational Linguistics.

\bibitem[{Lin et~al.(2017)Lin, Feng, dos Santos, Yu, Xiang, Zhou, and
  Bengio}]{LinFSYXZB17}
Zhouhan Lin, Minwei Feng, C{\'{\i}}cero~Nogueira dos Santos, Mo~Yu, Bing Xiang,
  Bowen Zhou, and Yoshua Bengio. 2017.
\newblock \href {https://openreview.net/forum?id=BJC\_jUqxe} {A structured
  self-attentive sentence embedding}.
\newblock In \emph{5th International Conference on Learning Representations,
  {ICLR} 2017, Toulon, France, April 24-26, 2017, Conference Track
  Proceedings}. OpenReview.net.

\bibitem[{Liu and Lane(2016)}]{LiuL16}
Bing Liu and Ian Lane. 2016.
\newblock \href {https://doi.org/10.21437/Interspeech.2016-1352}
  {Attention-based recurrent neural network models for joint intent detection
  and slot filling}.
\newblock In \emph{Interspeech 2016, 17th Annual Conference of the
  International Speech Communication Association, San Francisco, CA, USA,
  September 8-12, 2016}, pages 685--689. {ISCA}.

\bibitem[{Liu et~al.(2020{\natexlab{a}})Liu, Gu, Goyal, Li, Edunov,
  Ghazvininejad, Lewis, and Zettlemoyer}]{journals/corr/abs-2001-08210}
Yinhan Liu, Jiatao Gu, Naman Goyal, Xian Li, Sergey Edunov, Marjan
  Ghazvininejad, Mike Lewis, and Luke Zettlemoyer. 2020{\natexlab{a}}.
\newblock \href {http://arxiv.org/abs/2001.08210} {Multilingual denoising
  pre-training for neural machine translation}.
\newblock \emph{CoRR}, abs/2001.08210.

\bibitem[{Liu et~al.(2020{\natexlab{b}})Liu, Winata, Lin, Xu, and
  Fung}]{aaai/LiuWLXF20}
Zihan Liu, Genta~Indra Winata, Zhaojiang Lin, Peng Xu, and Pascale Fung.
  2020{\natexlab{b}}.
\newblock \href {https://aaai.org/ojs/index.php/AAAI/article/view/6362}
  {Attention-informed mixed-language training for zero-shot cross-lingual
  task-oriented dialogue systems}.
\newblock In \emph{The Thirty-Fourth {AAAI} Conference on Artificial
  Intelligence, {AAAI} 2020, The Thirty-Second Innovative Applications of
  Artificial Intelligence Conference, {IAAI} 2020, The Tenth {AAAI} Symposium
  on Educational Advances in Artificial Intelligence, {EAAI} 2020, New York,
  NY, USA, February 7-12, 2020}, pages 8433--8440. {AAAI} Press.

\bibitem[{Mesnil et~al.(2013)Mesnil, He, Deng, and
  Bengio}]{mesnil2013investigation}
Gr{\'e}goire Mesnil, Xiaodong He, Li~Deng, and Yoshua Bengio. 2013.
\newblock Investigation of recurrent-neural-network architectures and learning
  methods for spoken language understanding.
\newblock In \emph{INTERSPEECH}, pages 3771--3775.

\bibitem[{Mikolov et~al.(2013)Mikolov, Le, and Sutskever}]{MikolovLS13}
Tomas Mikolov, Quoc~V. Le, and Ilya Sutskever. 2013.
\newblock \href
  {http://dblp.uni-trier.de/db/journals/corr/corr1309.html#MikolovLS13}
  {Exploiting similarities among languages for machine translation.}
\newblock \emph{CoRR}, abs/1309.4168.

\bibitem[{Peters et~al.(2018)Peters, Neumann, Iyyer, Gardner, Clark, Lee, and
  Zettlemoyer}]{Peters2018}
Matthew~E. Peters, Mark Neumann, Mohit Iyyer, Matt Gardner, Christopher Clark,
  Kenton Lee, and Luke Zettlemoyer. 2018.
\newblock Deep contextualized word representations.
\newblock In \emph{Proc. of NAACL}.

\bibitem[{Pires et~al.(2019)Pires, Schlinger, and Garrette}]{PiresSG19}
Telmo Pires, Eva Schlinger, and Dan Garrette. 2019.
\newblock \href {http://dblp.uni-trier.de/db/conf/acl/acl2019-1.html#PiresSG19}
  {How multilingual is multilingual bert?}
\newblock In \emph{ACL (1)}, pages 4996--5001. Association for Computational
  Linguistics.

\bibitem[{Price(1990)}]{Price90}
P.~J. Price. 1990.
\newblock \href {http://dblp.uni-trier.de/db/conf/naacl/naacl1990.html#Price90}
  {Evaluation of spoken language systems: the atis domain.}
\newblock In \emph{HLT}. Morgan Kaufmann.

\bibitem[{Rongali et~al.(2020)Rongali, Soldaini, Monti, and
  Hamza}]{rongali2020don}
Subendhu Rongali, Luca Soldaini, Emilio Monti, and Wael Hamza. 2020.
\newblock Don't parse, generate! a sequence to sequence architecture for
  task-oriented semantic parsing.
\newblock \emph{arXiv preprint arXiv:2001.11458}.

\bibitem[{Ruder et~al.(2017)Ruder, Vulic, and Sogaard}]{ruder2017survey}
Sebastian Ruder, Ivan Vulic, and Anders Sogaard. 2017.
\newblock \href {http://arxiv.org/abs/1706.04902} {A survey of cross-lingual
  word embedding models}.
\newblock Cite arxiv:1706.04902.

\bibitem[{Schuster et~al.(2019{\natexlab{a}})Schuster, Gupta, Shah, and
  Lewis}]{schuster2018cross}
Sebastian Schuster, Sonal Gupta, Rushin Shah, and Mike Lewis.
  2019{\natexlab{a}}.
\newblock \href {https://doi.org/10.18653/v1/N19-1380} {Cross-lingual transfer
  learning for multilingual task oriented dialog}.
\newblock In \emph{Proceedings of the 2019 Conference of the North {A}merican
  Chapter of the Association for Computational Linguistics: Human Language
  Technologies, Volume 1 (Long and Short Papers)}, pages 3795--3805,
  Minneapolis, Minnesota. Association for Computational Linguistics.

\bibitem[{Schuster et~al.(2019{\natexlab{b}})Schuster, Ram, Barzilay, and
  Globerson}]{Schuster2019cross}
Tal Schuster, Ori Ram, Regina Barzilay, and Amir Globerson. 2019{\natexlab{b}}.
\newblock \href {https://www.aclweb.org/anthology/N19-1162} {Cross-lingual
  alignment of contextual word embeddings, with applications to zero-shot
  dependency parsing}.
\newblock In \emph{Proceedings of the 2019 Conference of the North {A}merican
  Chapter of the Association for Computational Linguistics: Human Language
  Technologies, Volume 1 (Long and Short Papers)}, pages 1599--1613,
  Minneapolis, Minnesota. Association for Computational Linguistics.

\bibitem[{Tang et~al.(2020)Tang, Tran, Li, Chen, Goyal, Chaudhary, Gu, and
  Fan}]{tang2020multilingual}
Yuqing Tang, Chau Tran, Xian Li, Peng-Jen Chen, Naman Goyal, Vishrav Chaudhary,
  Jiatao Gu, and Angela Fan. 2020.
\newblock Multilingual translation with extensible multilingual pretraining and
  finetuning.
\newblock \emph{arXiv preprint arXiv:2008.00401}.

\bibitem[{Tran et~al.(2020)Tran, Tang, Li, and Gu}]{tran2020cross}
Chau Tran, Yuqing Tang, Xian Li, and Jiatao Gu. 2020.
\newblock Cross-lingual retrieval for iterative self-supervised training.
\newblock \emph{arXiv preprint arXiv:2006.09526}.

\bibitem[{Upadhyay et~al.(2018)Upadhyay, Faruqui, Tür, Hakkani-Tür, and
  Heck}]{UpadhyayFTHH18}
Shyam Upadhyay, Manaal Faruqui, Gökhan Tür, Dilek Hakkani-Tür, and Larry~P.
  Heck. 2018.
\newblock \href
  {http://dblp.uni-trier.de/db/conf/icassp/icassp2018.html#UpadhyayFTHH18}
  {(almost) zero-shot cross-lingual spoken language understanding.}
\newblock In \emph{ICASSP}, pages 6034--6038. IEEE.

\bibitem[{Vaswani et~al.(2017)Vaswani, Shazeer, Parmar, Uszkoreit, Jones,
  Gomez, Kaiser, and Polosukhin}]{vaswani2017attention}
Ashish Vaswani, Noam Shazeer, Niki Parmar, Jakob Uszkoreit, Llion Jones,
  Aidan~N Gomez, {\L}ukasz Kaiser, and Illia Polosukhin. 2017.
\newblock Attention is all you need.
\newblock In \emph{Advances in neural information processing systems}, pages
  5998--6008.

\bibitem[{Williams et~al.(2016)Williams, Raux, and Henderson}]{WilliamsRH16a}
Jason~D. Williams, Antoine Raux, and Matthew Henderson. 2016.
\newblock \href
  {http://dblp.uni-trier.de/db/journals/dad/dad7.html#WilliamsRH16a} {The
  dialog state tracking challenge series: A review.}
\newblock \emph{D\&D}, 7(3):4--33.

\bibitem[{You et~al.(2019)You, Li, Reddi, Hseu, Kumar, Bhojanapalli, Song,
  Demmel, Keutzer, and Hsieh}]{you2019large}
Yang You, Jing Li, Sashank Reddi, Jonathan Hseu, Sanjiv Kumar, Srinadh
  Bhojanapalli, Xiaodan Song, James Demmel, Kurt Keutzer, and Cho-Jui Hsieh.
  2019.
\newblock Large batch optimization for deep learning: Training bert in 76
  minutes.
\newblock \emph{arXiv preprint arXiv:1904.00962}.

\bibitem[{Zhang and Wang(2016)}]{ZhangW16a}
Xiaodong Zhang and Houfeng Wang. 2016.
\newblock \href {http://www.ijcai.org/Abstract/16/425} {A joint model of intent
  determination and slot filling for spoken language understanding}.
\newblock In \emph{Proceedings of the Twenty-Fifth International Joint
  Conference on Artificial Intelligence, {IJCAI} 2016, New York, NY, USA, 9-15
  July 2016}, pages 2993--2999. {IJCAI/AAAI} Press.

\bibitem[{Zhu and Yu(2017)}]{ZhuY17}
Su~Zhu and Kai Yu. 2017.
\newblock \href
  {http://dblp.uni-trier.de/db/conf/icassp/icassp2017.html#ZhuY17}
  {Encoder-decoder with focus-mechanism for sequence labelling based spoken
  language understanding.}
\newblock In \emph{ICASSP}, pages 5675--5679. IEEE.

\end{thebibliography}
\bibliographystyle{acl_natbib}

\clearpage

\appendix
\insertFlatMoreResults
\section{Training Details}
\label{sec:appendix}
\paragraph{Settings for MTOP results in \cref{tab:1}}
For fine-tuning XLM-R, we use the Adam optimizer~\cite{kingma2014adam} with $\beta_1 = 0.9, \beta_2 = 0.99, \epsilon = 1e-6$ and batch size of 16.
We finetune for 20 epochs and search over learning rates $\in \{1,2,3\}e-5$ on dev set. All XLM-R models were run on single 32GB V100 Nvidia GPU.

For the XLU models in \cref{tab:1}, we use 300 dim XLU embeddings and feed them to a 2-layer 200 dim BiLSTM. The intent classification head contains an attention pooling layer as described in~\citet{LinFSYXZB17} with with attention dim 128 followed by a 200 dim linear projection before the softmax. The slot tagging head also contains a 200 dim linear layer followed by a CRF decoder. We use the we use the Adam optimizer with the same settings as above and a batch size of 32 for 40 epochs. The learning rate and BiLSTM dropouts are picked via a param sweep over the dev set.

\paragraph{Settings for MTOP results in \cref{tab:2}}
For training seq-2-seq models, we use stochastic weight averaging \citep{swa} with Lamb optimizer \citep{you2019large} and exponential learning rate decay for all models. For fine-tuning pre-trained models: we use batch size of 16 for all models except Marge, we use batch size 4 for Marge since we were not able to fit larger batch size into 32GB memory; We finetune for 50 epochs and again search over learning rates on dev set.

For copy pointer We use 1 layer multihead attention(MHA) with 4 attention heads to get copy distribution. For seq-2-seq model with XLM-R encoder, the decoder is a randomly initialized 3-layer transformer, with hidden size 1024 and 8 attention heads. XLM-R encoder (24 layers) is larger than mBART/CRISS/MARGE encoder (12 layers) so we were not able to fit a larger decoder into GPU memory.

For the XLU models specifically we use a 2-layer BiLSTM encoder with a hidden dimension of 256. For the decoder, we use a 2-layer LSTM with 256 dimension and a single attention head. Similar to the flat models, learning rate and LSTM dropouts are picked via a param sweep over the dev set. 

\paragraph{Settings for other benchmark results in \cref{tab:3}}
We use the same setting as described for \cref{tab:1} except for multilingual ATIS which doesn't have dev set, we just use the checkpoint after a fixed number of epochs.

\section{More Results}
\label{sec:more_results}
We report additional metrics for our experiments in this section. \cref{tab:7} contains the intent accuracy and slot F1 metrics of models for flat representation.

\end{document}